\documentclass[letterpaper, 10 pt, conference]{ieeeconf} 
\IEEEoverridecommandlockouts                            
\usepackage{algorithm}
\usepackage{algorithmic}
\usepackage{graphics} 
\usepackage{epsfig} 
\usepackage{times} 

\usepackage{graphicx}
\usepackage{array}
\usepackage{float}
\usepackage{afterpage}
\usepackage{dblfloatfix}

\usepackage[hidelinks]{hyperref}
\usepackage[symbol]{footmisc}
\usepackage{booktabs, multirow} 
\usepackage{soul}
\usepackage[table]{xcolor} 
\usepackage{changepage,threeparttable} 
\usepackage{amssymb}
\usepackage{amsmath}
\usepackage{amsfonts}
\usepackage{mathtools}
\usepackage{bbm}
\usepackage{cite}
\usepackage{microtype}
\AtBeginDocument{%
  \setlength{\abovedisplayskip}{4pt plus 2pt minus 2pt}%
  \setlength{\belowdisplayskip}{4pt plus 2pt minus 2pt}%
  \setlength{\abovedisplayshortskip}{2pt plus 1pt}%
  \setlength{\belowdisplayshortskip}{2pt plus 1pt}%
}
\title{\LARGE \bf Calibrated Keyframe-Segment Selection for Descriptor-Agnostic LiDAR Loop Closure in Repetitive Environments}

\author{Jaehyun Kim, Seungwon Choi, Wonseok Kang, and Tae-wan Kim%
\thanks{The authors are with the Department of Naval Architecture and Ocean Engineering, Seoul National University, Seoul, South Korea (e-mail: jaedalong@snu.ac.kr; csw3575@snu.ac.kr; eoid361@snu.ac.kr; taewan@snu.ac.kr). Corresponding author: Tae-wan Kim.}}

\hypersetup{
  pdftitle={Calibrated Keyframe-Segment Selection for Descriptor-Agnostic LiDAR Loop Closure in Repetitive Environments},
  pdfauthor={Jaehyun Kim, Seungwon Choi, Wonseok Kang, Tae-wan Kim},
  pdfsubject={QCSA preprint}
}

\begin{document}

\maketitle
\thispagestyle{plain}
\pagestyle{plain}


\begin{abstract}
False LiDAR loop closures can distort pose graphs in repetitive environments, where one query may match several distant places. Genuine revisits, unlike many perceptual aliases, tend to remain consistent across neighboring keyframes. We present Query-Calibrated Segmental Admission (QCSA), a loop-candidate selection method between descriptor retrieval and point-cloud registration. QCSA maps descriptor scores to evidence that distinguishes true loops from hard negatives, accumulates it over short keyframe segments, and admits only the best segment above a threshold fitted to the strongest false segment for each calibration query. The admitted segment forms two local submaps for one Generalized Iterative Closest Point (GICP) registration. QCSA requires pairwise descriptor scores and separate calibration for each descriptor and setting. Across seven descriptors on the SNU Library Dataset (SNULib), QCSA lowers average trajectory error with 53.8\% fewer registration attempts than Top-1. With common local submaps and downstream processing on NCLT, it also reduces registration attempts, constraints exceeding the pose-error tolerance, and trajectory error relative to Top-1. HeLiPR Bridge further evaluates repeated outdoor geometry. 

\end{abstract}

\section{Introduction}

\begin{figure*}
    \centering
    \includegraphics[width=1.0\linewidth]{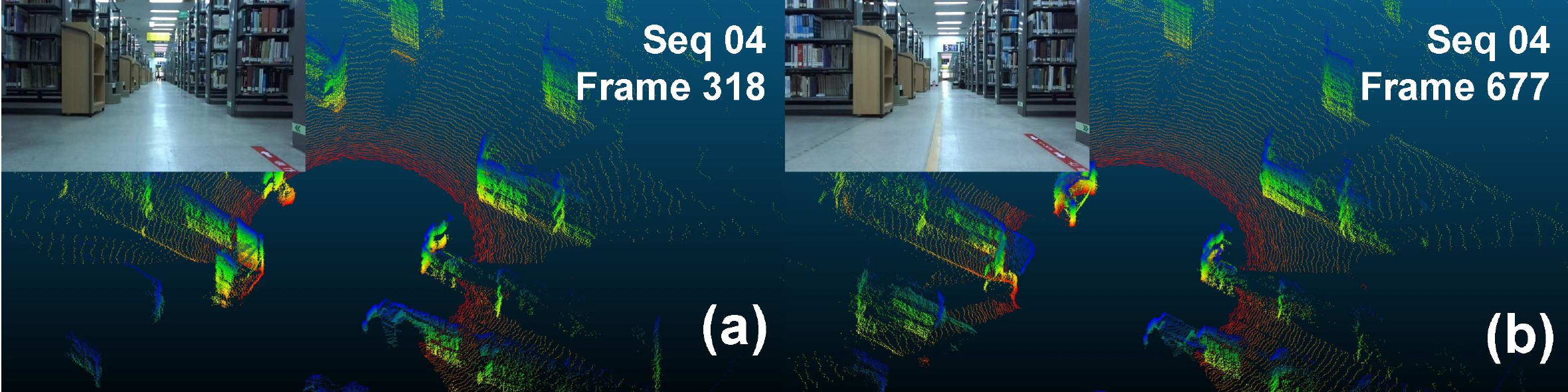}\\[0.6em]
    \includegraphics[width=1.0\linewidth]{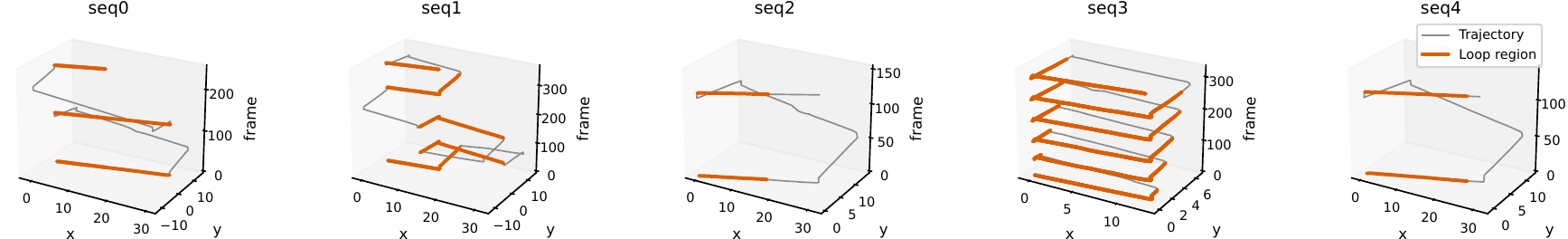}
    \caption{Repeated shelf geometry in SNULib at Seoul National University Library. Top: LiDAR scans and camera images at Sequence 04 keyframes 318 and 677, over 15\,m apart. Bottom: reference trajectories with revisited regions highlighted in orange.}
    \label{fig:example_of_data}
\end{figure*}

In LiDAR simultaneous localization and mapping (SLAM), loop closure retrieves a previously visited place and estimates its relative pose through point-cloud registration. The resulting constraint is added to pose-graph optimization (PGO) to correct accumulated drift~\cite{Dube2017SegMatch,Shan2020LIOSAM,Cadena2016}. Repetitive environments make this process unreliable. In Fig.~\ref{fig:example_of_data}, for example, locations separated by more than 15\,m contain nearly indistinguishable shelf geometry and can receive similarly strong descriptor scores. If such an aliased candidate also produces a locally plausible registration, the resulting constraint can distort the pose graph~\cite{Latif-RSS-12,Sunderhauf2012Switchable}. Selecting which retrieved candidates are worth registering is therefore critical.

Global descriptors provide efficient place retrieval~\cite{Kim2018ScanContext,Ma2022OverlapTransformer}, but repeated geometry can produce ambiguous descriptor matches~\cite{Gupta2026DensityMaps}. Sequence-based methods introduce temporal context through their representation or matching procedure~\cite{MilfordWyeth2012SeqSLAM,Vidanapathirana2021Locus,Ma2023SeqOT}, whereas geometric verification examines local correspondences or registers retrieved candidates~\cite{Rusu2009FPFH,Vidanapathirana2023SpectralGV,Laserna2025CliReg}. A practical gap remains between these stages: a candidate-selection rule that leaves an existing descriptor unchanged and determines, before registration, whether a retrieved match is supported by neighboring keyframes.

Our key observation is that a perceptual alias may be strong at one keyframe but often fails to remain coherent over neighboring motion-selected keyframes. A genuine revisit, in contrast, tends to retain positive evidence across neighboring keyframes. This suggests a simple test: instead of trusting the strongest match at one keyframe, require consistent evidence across a short keyframe segment.

Query-Calibrated Segmental Admission (QCSA) turns this test into a post-retrieval candidate-selection method. It maps each descriptor score to evidence that distinguishes a true loop from a top-ranked false match. QCSA accumulates this evidence over short paired keyframe segments and admits only the highest-scoring segment above a calibrated threshold. The threshold is fitted from the strongest false segment returned for each calibration query. The admitted segment forms two odometry-aligned local submaps for one Generalized Iterative Closest Point (GICP) registration~\cite{Segal2009GICP}. QCSA uses only scalar query--database scores; it leaves descriptor extraction and retrieval unchanged and can be used with different descriptors. Its score model and threshold are calibrated separately for each descriptor and setting.

We evaluate the complete front end with seven descriptors on SNULib and NCLT. Controlled NCLT and HeLiPR Bridge comparisons isolate candidate selection under common downstream processing in outdoor environments. On SNULib, QCSA's absolute trajectory error (ATE), evaluated as RMSE, averages $0.1188$\,m with 53.8\% fewer registration attempts than Top-1.

Our contributions are threefold. First, QCSA combines hard-negative score calibration, neighboring-keyframe consistency, and a query-calibrated threshold in a descriptor-agnostic candidate-selection method. Second, controlled comparisons with common local-submap construction and downstream processing isolate candidate selection. Third, loop-closure experiments cover seven descriptors on our multi-traversal SNULib dataset and NCLT, plus repeated outdoor Bridge sequences.

\section{Related Work}

\subsection{LiDAR Place Retrieval}
LiDAR place retrieval represents a keyframe or local map with a descriptor and ranks database observations by similarity. Global descriptor design has been studied extensively through handcrafted projection and distribution representations~\cite{Kim2018ScanContext,Kim2022ScanContext++,Wang2020LiDAR-Iris,Xu2023RING++,Liao2024NDTMapCode} and learned point-cloud embeddings~\cite{Ma2022OverlapTransformer,Vidanapathirana2022LoGG3DNet,Komorowski2022MinkLoc3Dv2}. Although these approaches differ in representation, training, and score distribution, they expose ranked candidates and scalar scores. QCSA operates on this common output and leaves descriptor extraction and retrieval unchanged.

\subsection{Temporal and Calibrated Matching}
Temporal context may be incorporated into the representation or the matching procedure. SeqSLAM~\cite{MilfordWyeth2012SeqSLAM} matches image sequences under appearance change. Locus~\cite{Vidanapathirana2021Locus} pools topological and temporal scene information into a global LiDAR descriptor, while SeqOT~\cite{Ma2023SeqOT} learns a descriptor from sequential LiDAR range images. QCSA retains the original single-keyframe descriptor and evaluates short paired segments anchored at its Top-$M$ retrievals.

Probabilistic place recognition, including FAB-MAP~\cite{Cummins2008FABMAP}, models the likelihood of an observation at a place. Vysotska et al.~\cite{Vysotska2025AdaptiveThresholding} update an image-similarity threshold online from a sliding patch of the similarity matrix without labeled calibration data. QCSA uses labeled calibration trajectories to map an existing descriptor's scores to true-loop versus top-ranked hard-negative evidence. It then fits one admission threshold to the highest-scoring segment containing no true-loop pair for each calibration query.

\subsection{Candidate Verification and Loop-Closure Front Ends}
Geometric verification evaluates retrieved candidates from local correspondences. Matching Fast Point Feature Histogram (FPFH) descriptors~\cite{Rusu2009FPFH} yields local correspondences, from which random sample consensus (RANSAC)~\cite{FischlerBolles1981} estimates a rigid transformation. SpectralGV~\cite{Vidanapathirana2023SpectralGV} computes a registration-free geometric consistency score, and CliReg~\cite{Laserna2025CliReg} uses a maximal clique of compatible correspondences for loop validation. These methods require geometric features for the candidates under consideration. QCSA instead uses descriptor-score consistency to select a segment before registration and sends only the admitted segment to GICP.

Denniston et al.~\cite{Denniston2022Prioritization} prioritize loop closures using pose-graph information, point-cloud observability, and beacon proximity in multi-robot SLAM\@. MapClosures~\cite{Gupta2026DensityMaps} builds odometry-based local maps and prunes self-similar features from bird's-eye-view density images. QCSA instead selects short segments from an existing descriptor's scalar scores before registration; the admitted segment supplies the local submaps.

\section{Method}
\label{sec:method}

\begin{figure*}[t]
    \centering
    \includegraphics[width=0.99\linewidth]{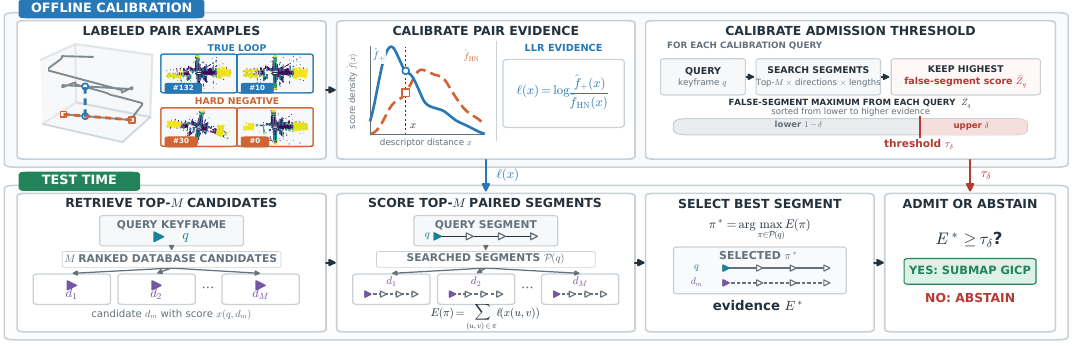}
\caption{QCSA data flow. Pair calibration converts retrieval scores to loop evidence, and query-level calibration sets $\tau_\delta$ from the highest-scoring all-false segment for each calibration query. At test time, Top-$M$ candidates anchor paired segments; QCSA admits the best segment above $\tau_\delta$ and applies one local-submap GICP registration.}
    \label{fig:overview}
\end{figure*}
  
\subsection{Overview}
\label{subsec:qcs_overview}
QCSA operates between global descriptor retrieval and GICP registration. The SLAM front end supplies keyframe clouds and odometry poses for $N_q$ motion-selected query keyframes and $N_d$ database keyframes. For query index $0\le q<N_q$, retrieval returns $\mathcal C(q)=\{d_1,\ldots,d_M\}$. Here, $0\le d_m<N_d$ is the index of the $m$th-ranked database candidate, and $x(q,d_m)$ is its descriptor score. Scores between neighboring keyframe pairs are computed from stored descriptors and reused across segment lengths.

Retrieval provides the Top-$M$ candidates. QCSA selects the highest-scoring paired segment and sends it to GICP only when its score reaches the calibrated threshold. GICP then estimates the relative pose between the admitted local submaps, and a registration satisfying the quality thresholds is inserted as a pose-graph constraint (Fig.~\ref{fig:overview}). Algorithm~\ref{alg:qcsa_s2s} summarizes the procedure.

Pair-level calibration distinguishes true-loop scores from highly ranked false matches. Segment scoring then tests whether this evidence persists over neighboring keyframes. Finally, query-level calibration sets the threshold applied after searching all candidates, directions, and segment lengths.

\subsection{Pair-Level Calibration}
\label{subsec:qcs_llr}
Pair-level calibration compares the retrieval score of each candidate pair with the score distributions of true loops and top-ranked false matches. For each descriptor, calibration trajectories provide score distributions for true loops and false pairs among the $M$ highest-ranked candidates. We call the latter hard-negative (HN) pairs: the incorrect matches most likely to be retrieved. They model the descriptor-specific aliases that directly compete with true loops after retrieval.

We estimate each one-dimensional distribution with Gaussian kernel density estimation. For a class with $n$ samples and score standard deviation $\hat\sigma$, we use Silverman's Gaussian rule-of-thumb bandwidth $1.06\hat\sigma n^{-1/5}$~\cite{Silverman1986}.

Given a descriptor score $x$, we compute a clipped log-likelihood ratio (LLR)
\begin{equation}
\label{eq:llr}
\ell(x)
=
\mathrm{clip}_{[\ell_{\min},\ell_{\max}]}
\left(
\log \frac{\hat f_{+}(x)}{\hat f_{\mathrm{HN}}(x)}
\right),
\end{equation}
where $\hat f_{+}$ and $\hat f_{\mathrm{HN}}$ are the true-loop and hard-negative score densities, and $[\ell_{\min},\ell_{\max}]$ is the clipping interval. A positive $\ell(x)$ favors a true loop, whereas a negative value favors a hard negative. We clip the LLR so that one score in a range containing few calibration samples cannot outweigh the remaining keyframes in a segment.

\subsection{Segment Evidence over Neighboring Keyframes}
\label{subsec:qcs_segments}
Segment scoring tests whether the pair-level evidence persists across neighboring keyframes. QCSA operates on successive keyframes selected by the SLAM front end using motion thresholds, together with their clouds and odometry poses. For an integer segment length $h\in\{n_{\min},\ldots,n_{\max}\}$, each retrieved database keyframe $d_m$ anchors the ordered sequence of query--database pairs
\begin{equation}
\label{eq:qcs_path}
\pi(q,d_m;s,h)
=
\bigl((q+i,\; d_m+s i)\bigr)_{i=0}^{h-1},
\end{equation}
where $s\in\{+1,-1\}$ sets the database traversal direction. A segment is valid if $q+h\le N_q$ and $0\le d_m+s i<N_d$ for every $i$. The paired segments are expected to span similar travel distances.

In a genuine revisit, neighboring keyframes are expected to form a coherent path of positive evidence. A repeated-scene alias can produce a high anchor score, but its segment score decreases when that evidence does not persist. The two directions allow either traversal order. Because keyframes are selected by motion, advancing by one keyframe approximates equal-distance rather than equal-time progress.

For each pair $(u,v)\in\pi$, $u$ and $v$ index the query and database keyframes, respectively. The segment evidence is the sum of its per-pair LLRs:
\begin{equation}
\label{eq:qcs_segment_score}
E(\pi)
=
\sum_{(u,v)\in\pi} \ell(x(u,v)).
\end{equation}
Because adjacent scores are correlated, we treat $E(\pi)$ as an accumulated evidence score. Its unnormalized sum favors evidence that persists over more keyframes.
Let $\mathcal{P}(q)$ contain every paired segment formed from the $M$ retrieved candidates, the two database directions, and the allowed segment lengths. The method selects the segment with the largest sum:
\begin{equation}
\label{eq:qcs_best_segment}
\pi^*(q)=\operatorname*{arg\,max}_{\pi\in\mathcal{P}(q)}E(\pi),
\qquad
E^*(q)=E(\pi^*(q)).
\end{equation}
The selected segment is admitted to registration only when $E^*(q)$ reaches the threshold defined next. Calibration and inference use the same candidate count, directions, and segment-length range.

\subsection{Query-Level Admission Threshold}
\label{subsec:qcs_calibration}
Segment scoring identifies the strongest candidate, but it does not determine whether that candidate is reliable enough to register. We set this admission threshold offline by applying the same segment search to calibration queries.

For each calibration query $q$, let $\mathcal P_0(q)$ contain the candidate segments with no reference loop pair. Among them, we retain only the highest score:
\begin{equation}
\label{eq:qcs_false_max}
\widetilde Z_q=\max_{\pi\in\mathcal P_0(q)}E(\pi).
\end{equation}
The value $\widetilde Z_q$ is therefore the most convincing false segment that the method could select for query $q$. If no such segment exists, we set $\widetilde Z_q=-\infty$. Repeating this procedure gives one false-segment maximum per calibration query. This matches the test-time search over candidates, directions, and lengths, and prevents queries that produce more candidate segments from receiving greater weight.

For an upper-tail fraction $\delta\in(0,1)$, we sort these maxima and use the following finite-sample rank rule to select the corresponding quantile~\cite{angelopoulos2023conformal}:
\begin{equation}
\label{eq:qcs_boundary}
k_\delta
=
\min\!\left\{n_{\mathrm{cal}},\left\lceil (n_{\mathrm{cal}}+1)(1-\delta) \right\rceil\right\},\qquad
\tau_\delta = \widetilde Z_{(k_\delta)},
\end{equation}
where $n_{\mathrm{cal}}$ is the number of calibration queries and $\widetilde Z_{(1)}\le\cdots\le \widetilde Z_{(n_{\mathrm{cal}})}$ are the sorted maxima. A smaller $\delta$ selects a higher, more conservative threshold; a larger $\delta$ sends more segments to registration. At test time, QCSA admits the best segment when $E^*(q)\ge \tau_\delta$. If $\tau_\delta$ is non-finite, the calibration data do not support the selected quantile, and QCSA admits no candidate for that fold.

\subsection{Local-Submap Registration and Loop Insertion}
\label{subsec:qcs_insertion}
An admitted segment determines both the point clouds used for registration and the two keyframes connected by the resulting loop constraint. Let $j_c=\lfloor(|\pi^*(q)|-1)/2\rfloor$ and $(q_c,d_c)=\pi^*(q)[j_c]$, using the lower-indexed center pair for an even segment length. GICP estimates the relative pose between $q_c$ and $d_c$; the other keyframes in the segment provide additional point-cloud support for that estimate.

Let $\mathcal X_i$ be the point cloud in keyframe $i$, and let $\mathbf T_i\in\mathrm{SE}(3)$ map that keyframe into the odometry frame before loop correction. We downsample each cloud at resolution $v_p$ and transform it into the corresponding center frame. We then merge the transformed clouds on each side and downsample each submap at resolution $v_s$:
\begin{align}
\mathcal S_q(\pi)&=\mathcal V_{v_s}\!\left(\bigcup_{(u,v)\in\pi}\mathbf T_{q_c}^{-1}\mathbf T_u\mathcal V_{v_p}(\mathcal X_u)\right),\label{eq:query_submap}\\
\mathcal S_d(\pi)&=\mathcal V_{v_s}\!\left(\bigcup_{(u,v)\in\pi}\mathbf T_{d_c}^{-1}\mathbf T_v\mathcal V_{v_p}(\mathcal X_v)\right),
\label{eq:database_submap}
\end{align}
where $\mathcal V_{\nu}$ denotes voxel-grid downsampling at resolution $\nu$. One coarse-to-fine GICP registration aligns $\mathcal S_d$ to $\mathcal S_q$ and estimates $\hat{\mathbf T}_{q_c d_c}\in\mathrm{SE}(3)$, which maps points from the database-center frame $d_c$ to the query-center frame $q_c$.

We accept the registration using two geometric checks. Let $\rho$ be the fraction of aligned database-submap points with a query-submap neighbor within $r_{\mathrm{in}}$, and let $\varepsilon$ be the root-mean-square error (RMSE) of those correspondences. If $\rho\ge\rho_{\min}$ and $\varepsilon\le\varepsilon_{\max}$, the relative pose is inserted as one factor between $q_c$ and $d_c$.

Different segment hypotheses may cover the same portion of the query trajectory and otherwise produce redundant loop constraints. Let $\mathcal I_q(\pi)$ be the contiguous set of query-keyframe indices in segment $\pi$. QCSA sends a segment to registration only when this interval does not overlap one stored in $\mathcal R$. After a successful registration, its interval is added to $\mathcal R$; a failed registration leaves the interval available. Thus, each admitted segment triggers one GICP registration and yields at most one loop constraint. Because a segment is processed after its last query keyframe arrives, its delay is at most $n_{\max}-1$ keyframes.

\begin{algorithm}[t]
\caption{QCSA Admission and Loop-Constraint Generation}
\label{alg:qcsa_s2s}
\footnotesize
\begin{algorithmic}[1]
\REQUIRE Admission inputs: candidate sets $\{\mathcal C(q)\}$; pair scores $x$; calibrated $\ell$ and $\tau_\delta$; segment lengths $n_{\min},n_{\max}$.
\REQUIRE Registration inputs: point clouds $\{\mathcal X_i\}$; odometry poses $\{\mathbf T_i\}$; parameters $v_p,v_s,r_{\mathrm{in}},\rho_{\min},\varepsilon_{\max}$.
\ENSURE Inserted loop-constraint set $\mathcal F$.
\STATE $\mathcal R\leftarrow\emptyset$; $\mathcal F\leftarrow\emptyset$
\IF{$\tau_\delta\notin\mathbb R$}
    \RETURN $\mathcal F$
\ENDIF
\FORALL{queries $q$ in temporal order}
    \STATE $\mathcal P(q)\leftarrow\{\pi(q,d_m;s,h):d_m\in\mathcal C(q),\ s\in\{-1,+1\},\ h\in\{n_{\min},\ldots,n_{\max}\},\ \pi\text{ valid}\}$
    \IF{$\mathcal P(q)\neq\emptyset$}
        \STATE $\pi^*\leftarrow\arg\max_{\pi\in\mathcal P(q)}E(\pi)$; $E^*\leftarrow E(\pi^*)$
        \IF{$E^*\ge \tau_\delta$ and $\mathcal I_q(\pi^*)\cap\bigcup\mathcal R=\emptyset$}
            \STATE $j_c\leftarrow\lfloor(|\pi^*|-1)/2\rfloor$; $(q_c,d_c)\leftarrow\pi^*[j_c]$
            \STATE $(\mathcal S_q^*,\mathcal S_d^*)\leftarrow(\mathcal S_q(\pi^*),\mathcal S_d(\pi^*))$ \hfill{\scriptsize Eqs.~\eqref{eq:query_submap}--\eqref{eq:database_submap}}
            \STATE $(\hat{\mathbf T}_{q_c d_c},\rho,\varepsilon)\leftarrow\operatorname{GICP}(\mathcal S_d^*,\mathcal S_q^*)$
            \IF{$\rho\ge\rho_{\min}$ and $\varepsilon\le\varepsilon_{\max}$}
                \STATE $\mathcal R\leftarrow\mathcal R\cup\{\mathcal I_q(\pi^*)\}$
                \STATE $\mathcal F\leftarrow\mathcal F\cup\{(q_c,d_c,\hat{\mathbf T}_{q_c d_c})\}$
            \ENDIF
        \ENDIF
    \ENDIF
\ENDFOR
\RETURN $\mathcal F$
\end{algorithmic}
\end{algorithm}

Calibration trajectories provide $\hat f_{+}$, $\hat f_{\mathrm{HN}}$, and $\tau_\delta$ before a test trajectory is processed. A new sensor, environment, preprocessing method, or descriptor requires these quantities to be refitted.

\section{Experimental Setup}
\label{sec:dataset_and_eval}

We evaluate QCSA in two settings. The controlled comparison uses the same centered 11-keyframe submaps, GICP settings, overlapping-segment suppression, and pose graph for every method; only the candidate sent to registration differs. This comparison isolates candidate selection. The complete front-end evaluation instead builds both local submaps from the segment selected by QCSA and evaluates candidate selection and registration together.

\subsection{Datasets}

The SNU Library Dataset (SNULib) contains five sequences through nearly identical shelf rows and corridors in the book repository at Seoul National University Library. We recorded them with a Clearpath Jackal carrying an Ouster OS1-128 LiDAR, an inertial measurement unit, and a stereo camera. The SLAM front end selects keyframes by motion; consecutive keyframes are 0.56--0.81\,m apart across sequences and approximately 0.7\,m apart on average. Keyframe clouds and odometry poses are fixed across methods. Reference trajectories are obtained with the A-LOAM LiDAR odometry and mapping system~\cite{Qin2018ALOAM,zhang2014loam}, followed by hierarchical LiDAR bundle adjustment (HBA)~\cite{HBA}. Our implementation and dataset will be released at \url{https://github.com/wanderingcar/snu_library_dataset}.

We also use two Bridge sequence pairs from HeLiPR~\cite{Jung2024HeLiPR} and both directions of three NCLT~\cite{nclt} session pairs. These provide repeated outdoor bridge structures and six directed evaluations among the NCLT 2012-01-08, 2012-01-15, and 2012-02-02 sessions.

\begin{table}[!b]
\centering
\caption{Experimental parameters and their scope.}
\label{tab:hyperparams}
\setlength{\tabcolsep}{2.3pt}
\scriptsize
\begin{tabular}{@{}lll@{}}
\toprule
Parameter & Value & Scope \\
\midrule
Candidates per query & $M=4$ & all Top-4 methods \\
Temporal exclusion & 25 keyframes & SNULib only \\
Segment length & $h\in\{5,\ldots,11\}$ & QCSA \\
Database direction & $s\in\{+1,-1\}$ & QCSA \\
LLR range & $[-4,4]$ & QCSA \\
Threshold setting & $\delta=0.30$ & QCSA \\
$v_p/v_s/r_{\mathrm{in}}$ & $0.30/0.25/0.50$\,m & all datasets \\
Indoor registration & inlier ratio$\geq0.70$, RMSE$\leq0.15$\,m & SNULib \\
NCLT registration & inlier ratio$\geq0.70$, RMSE$\leq0.20$\,m & NCLT \\
Bridge registration setting & selected on the other pair & Bridge \\
\bottomrule
\end{tabular}
\end{table}

\subsection{Evaluation Procedure}
Top-1, Top-4 re-ranking, and QCSA share the same retrieval scores and candidates; sequence and learned local-feature methods retain their own matching procedures. Unless stated otherwise, all candidates use the common GICP registration and pose graph described below.

\subsubsection{Calibration and test splits}
\label{subsec:eval_protocol_lib}

Calibration and test data are separated by trajectory or session. The library data use leave-one-sequence-out calibration. Bridge alternates its two session pairs: both directions of one pair calibrate both directions of the other. NCLT holds out one query session, calibrates on both directions between the remaining sessions, and tests its two directed query$\to$database pairs. A descriptor fold contains 880--1,096 library, 115--128 Bridge, or 1,230--1,348 NCLT calibration queries. Calibration is completed before the held-out trajectory, whose reference poses are used only for evaluation.

We choose $M=4$, $h\in\{5,\ldots,11\}$, LLR clipping to $[-4,4]$, and $\delta=0.30$ on SNULib and retain them elsewhere (Table~\ref{tab:hyperparams}). Each descriptor and setting refits its score densities and admission threshold. The densities use a 256-point grid, linear interpolation, and a $10^{-12}$ floor.

Within-session library retrieval excludes the 25 most recent database keyframes; Bridge and NCLT use cross-session retrieval. For both outdoor datasets, query and database anchors are spaced by 10\,m and 5\,m, respectively, following HeLiPR~\cite{Jung2024HeLiPR}, while segment scoring uses 1\,m keyframe streams in both directions.

\subsubsection{Reference loop definitions}
Spatial proximity alone does not distinguish repeated shelving in SNULib, so reference loops use point-cloud overlap. For temporally separated pairs within $5$\,m, reference-pose-aligned clouds are voxelized at $0.20$\,m. Directed overlap is the fraction of points with a neighbor in the other cloud within $0.30$\,m; a pair is positive when the bidirectional mean is at least $0.30$. A positive segment contains at least five positive pairs, matching $n_{\min}$, with gaps of at most two. Segments with one to four positives are ambiguous and excluded from threshold fitting. Of the 16,652 positive pairs, 1,250 have opposite headings, but none forms a positive segment.

For Bridge and NCLT calibration, pairs within $7.5$\,m are positive, pairs beyond $20$\,m are negative, and intermediate pairs are excluded.

\subsubsection{Descriptors and baselines}
We evaluate LiDAR Iris~\cite{Wang2020LiDAR-Iris}, ScanContext++~\cite{Kim2022ScanContext++}, NDT-Map-Code (NDT-MC)~\cite{Liao2024NDTMapCode}, OverlapTransformer (OT)~\cite{Ma2022OverlapTransformer}, LoGG3D-Net~\cite{Vidanapathirana2022LoGG3DNet}, RING++~\cite{Xu2023RING++}, and MinkLoc3Dv2~\cite{Komorowski2022MinkLoc3Dv2}. Each descriptor is evaluated separately with its own retrieval scores and QCSA calibration. Learned descriptors use published pretrained weights, and descriptor- and dataset-specific preprocessing is fixed before calibration.

We use Top-1 with single-scan GICP~\cite{Segal2009GICP} as a pairwise baseline. Two Top-4 baselines re-rank FPFH correspondences~\cite{Rusu2009FPFH} using either RANSAC~\cite{FischlerBolles1981} or SpectralGV~\cite{Vidanapathirana2023SpectralGV}. Both use 0.5\,m voxels, 1.0/2.5\,m normal/feature radii, and at most 128 deterministic voxel-cloud keypoints. RANSAC uses mutual matching, three-point hypotheses, a 0.8 edge-length check, 10,000 iterations, and 0.999 confidence; inlier fitness re-ranks the candidates. Its correspondence limit is 1.0\,m on the library and Bridge data and 0.5\,m on NCLT. SpectralGV uses a 0.4\,m consistency threshold and five power iterations. The sequence baselines are SeqSLAM~\cite{MilfordWyeth2012SeqSLAM} and SeqOT~\cite{Ma2023SeqOT}. We also evaluate intermediate MinkLoc3Dv2~\cite{Komorowski2022MinkLoc3Dv2} features with RANSAC or SpectralGV\@. FPFH provides a common local representation for the seven descriptors on NCLT, whereas MapClosures~\cite{Gupta2026DensityMaps} retains its local-map construction and registration procedure.

The controlled NCLT comparison constructs centered 11-keyframe submaps for Top-1, raw-score segments, hard-negative LLR with a zero threshold, QCSA, and Top-4 FPFH-RANSAC\@. Retrieval outputs a distance score for every descriptor, with smaller values indicating stronger matches. Using the interquartile range (IQR) of the calibration scores, the raw-score method maps a retrieval distance $x$ to
\begin{equation}
\begin{aligned}
\ell_{\mathrm{raw}}(x)
&=\mathrm{clip}_{[-4,4]}
\left(
\frac{\mathrm{median}(X_{\mathrm{cal}})-x}{\sigma_{\mathrm{IQR}}}
\right),\\
\sigma_{\mathrm{IQR}}
&=\frac{Q_{0.75}(X_{\mathrm{cal}})-Q_{0.25}(X_{\mathrm{cal}})}{1.349}.
\end{aligned}
\label{eq:raw_control}
\end{equation}
where $X_{\mathrm{cal}}$ contains the two calibration directions' Top-4 distances and $Q_p$ is its $p$th quantile. Since 1.349 is the standard-normal IQR, the division gives a robust standard-deviation scale without assuming Gaussian scores. If the IQR is zero, we use the standard deviation; if both vanish, a unit scale prevents division by zero. Raw-score segments replace the hard-negative LLR with $\ell_{\mathrm{raw}}$ while retaining QCSA's segment search, clipping, and a separately fitted false-segment threshold. The zero-threshold baseline retains the hard-negative LLR and registers segments satisfying $E^*(q)\geq0$. All methods share registration initialization, GICP, overlapping-segment suppression, graph optimization, and the centered $n_{\max}=11$ submap window.

\subsubsection{Evaluation Metrics and Settings}
We report loop-constraint count, place and relative-pose precision, and ATE, with ATE expressed as RMSE\@. Place precision uses positive overlap on SNULib and a center distance within $7.5$\,m outdoors. Relative-pose precision additionally requires $0.5$\,m/$5^\circ$; Bridge also reports $1$\,m/$5^\circ$. These tolerances define the evaluation metrics, while GICP acceptance uses the limits in Table~\ref{tab:hyperparams}. The NCLT 95th percentile is taken over the 42 descriptor--direction ATE values. A registration attempt is one candidate submitted to initialization and GICP. For Table~\ref{tab:nclt}(b), 0.25\,m odometry groups are evaluation units, and a group pair is positive if it contains an overlap-positive reference pair.

SNULib trajectories are aligned to the reference at the first pose in SE(2). On NCLT and Bridge, the first inserted loop places the database odometry in the query frame, and only the query root is fixed during PGO\@. NCLT uses one SE(3) alignment for the combined two-session graph, whereas Bridge uses one SE(2) alignment; both are applied only to compute ATE\@. Each graph is optimized once.

Except for MapClosures, all methods use the same GICP~\cite{Segal2009GICP}. The coarse stage uses a 0.5\,m voxel grid, a 1.5\,m correspondence limit, and 30 iterations; the fine stage uses the 0.25\,m submaps, a 0.5\,m limit, and 50 iterations. On NCLT, a ground-truth-free $360^\circ$ yaw search between the center scans initializes the common GICP refinement. On Bridge, Top-1 and QCSA use the anchor keyframes' relative odometry yaw, whereas FPFH-RANSAC supplies its estimated transformation.

All methods use a common Gaussian pose graph without a robust loss so that ATE reflects front-end constraint quality. The noise standard deviations are 0.05\,m/$2^\circ$ for odometry and 0.10\,m/$5^\circ$ for NCLT and SNULib loop constraints. NCLT uses the fixed outdoor RMSE limit of 0.20\,m. Bridge considers RMSE limits of 0.1875--0.25\,m and loop standard deviations of 0.5--1.0\,m/$1$--$5^\circ$; the setting with the lowest mean calibration-pair ATE among those initializing every graph is fixed for the held-out pair and shared by all methods.

The inlier ratio is the fraction of source points with a target neighbor within the final correspondence distance. RMSE is computed over those correspondences. With $M=4$ and $n_{\max}=11$, QCSA accesses at most 84 unique pair scores per query, including the four anchor scores.

\section{Experimental Results}
\label{sec:experiments}

\subsection{Results on SNULib}
On SNULib, 199 of 202 inserted constraints meet the $0.5$\,m/$5^\circ$ relative-pose threshold across 35 descriptor--trajectory evaluations. QCSA's ATE RMSE averages $0.1188$\,m, compared with $0.1238$\,m for Top-1 and $0.1194$\,m for FPFH-RANSAC, while requiring 53.8\% fewer registration attempts than Top-1 (Table~\ref{tab:library_efficiency}). QCSA yields lower ATE than FPFH-RANSAC in 17 evaluations, FPFH-RANSAC in 16, and two are tied. On Sequence 02, no QCSA candidate passes geometric verification; hence no loop constraint is inserted, and all descriptors retain the odometry-only ATE of $0.0980$\,m.

\begin{figure}[!b]
    \centering
    \includegraphics[width=0.96\linewidth]{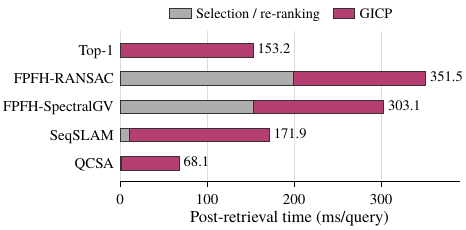}
    \caption{Post-retrieval time for 8,687 SNULib queries. Bars show candidate selection or re-ranking and GICP. Totals exclude descriptor extraction, score-cache construction, and PGO; FPFH extraction is included.}
    \label{fig:library_runtime}
\end{figure}

Figure~\ref{fig:library_runtime} separates candidate selection or re-ranking from GICP on an AMD Ryzen 7 5800X CPU\@. With descriptor distances precomputed, QCSA requires 68.1\,ms/query, 55.5\% less than Top-1; its 1.3\,ms selection cost is small relative to the registration time saved. Pair-score cache construction and PGO are excluded; FPFH extraction is included in re-ranking.

\begin{table}[!b]
\centering
\caption{Loop-closure and trajectory results on SNULib over 35 descriptor--trajectory evaluations.}
\label{tab:library_efficiency}
\setlength{\tabcolsep}{3.0pt}
\scriptsize
\begin{tabular}{@{}lrrr@{}}
\toprule
\multicolumn{4}{l}{(a) Aggregate trajectory accuracy} \\
\addlinespace[1pt]
Method & Reg./query & Constraints & ATE mean/max. (m) \\
\midrule
Top-1 & 0.895 & 954 & 0.1238 / 0.1851 \\
Top-4 FPFH-RANSAC & 0.895 & 958 & 0.1194 / \textbf{0.1638} \\
Top-4 FPFH-SpectralGV & 0.895 & 1,010 & 0.1237 / 0.1718 \\
SeqSLAM & 0.844 & 573 & 0.2627 / 2.1959 \\
\textbf{QCSA + segment submaps} & \textbf{0.414} & 202 & \textbf{0.1188} / 0.2091 \\
\bottomrule
\end{tabular}
\vspace{1pt}

{\fontsize{6}{6.3}\selectfont
\setlength{\tabcolsep}{1.45pt}
\begin{tabular}{@{}lrrrrr@{}}
\toprule
\multicolumn{6}{l}{(b) Sequence-wise ATE RMSE (m); entries are QCSA/FPFH-RANSAC} \\
\addlinespace[1pt]
Descriptor & 00 & 01 & 02 & 03 & 04 \\
\midrule
LiDAR Iris & 0.112/0.126 & 0.167/0.146 & 0.098/0.073 & 0.111/0.125 & 0.113/0.162 \\
ScanContext++ & 0.111/0.107 & 0.186/0.120 & 0.098/0.081 & 0.113/0.110 & 0.113/0.161 \\
NDT-MC & 0.103/0.078 & 0.154/0.159 & 0.098/0.088 & 0.110/0.136 & 0.114/0.163 \\
OT & 0.117/0.107 & 0.169/0.142 & 0.098/0.098 & 0.119/0.136 & 0.113/0.109 \\
LoGG3D-Net & 0.117/0.113 & 0.103/0.114 & 0.098/0.098 & 0.108/0.109 & 0.113/0.116 \\
RING++ & 0.108/0.111 & 0.209/0.127 & 0.098/0.087 & 0.112/0.129 & 0.113/0.164 \\
MinkLoc3Dv2 & 0.110/0.070 & 0.123/0.159 & 0.098/0.075 & 0.116/0.122 & 0.116/0.160 \\
\bottomrule
\end{tabular}}
\end{table}

The SNULib ablation separates the registration input from candidate selection. With QCSA candidates, single-scan GICP gives $0.1224$\,m mean ATE and 96.17\% relative-pose precision, whereas segment submaps give $0.1188$\,m and 98.51\%. Candidate-selection effects are evaluated separately under common submap construction in Table~\ref{tab:nclt_matched}. On the five SNULib trajectories, MinkLoc3Dv2-feature RANSAC and SpectralGV give $0.1271$ and $0.1164$\,m mean ATE, compared with $0.1128$\,m for QCSA using MinkLoc3Dv2; SeqOT gives $0.1830$\,m.

\subsection{Controlled Candidate Selection on NCLT}
On NCLT, every method in Table~\ref{tab:nclt_matched} uses the same centered 11-keyframe submap construction and downstream processing; only candidate selection differs. We fix the outdoor registration limit at 0.20\,m. Each descriptor is calibrated on the other sessions before evaluation on the held-out query session.

\begin{table*}[!t]
\centering
\caption{Controlled NCLT comparison over 42 descriptor--route configurations with common 11-keyframe submaps, GICP, duplicate suppression, and pose graph. Pose-error denotes constraints exceeding $0.5$\,m/$5^\circ$.}
\label{tab:nclt_matched}
\setlength{\tabcolsep}{3.1pt}
\scriptsize
\begin{tabular}{@{}lcccrrrrrr@{}}
\toprule
\multicolumn{10}{l}{(a) Aggregate admission and trajectory results} \\
\addlinespace[1pt]
& \multicolumn{3}{c}{Admission components} & \multicolumn{6}{c}{} \\
\cmidrule(lr){2-4}
Admission rule & HN LLR & Segment & Calib. $\tau_\delta$ & Reg.\ attempts & Constraints & Pose-error & Pose precision & ATE mean/95th (m) & ATE $>10$\,m \\
\midrule
Top-1 + 11-keyframe submaps & -- & -- & -- & 25,668 & 1,629 & 48 & 97.05\% & 6.223/11.542 & 7 \\
Top-4 FPFH-RANSAC & -- & -- & -- & 25,661 & 1,635 & 52 & 96.82\% & 9.199/27.555 & 8 \\
\midrule
Raw-score segments & -- & \checkmark & \checkmark & 19,206 & 1,652 & 36 & 97.82\% & 5.877/14.601 & 3 \\
HN-LLR segment, zero threshold & \checkmark & \checkmark & -- & \textbf{16,999} & 1,518 & \textbf{27} & \textbf{98.22\%} & 5.618/9.439 & \textbf{2} \\
\midrule
\rowcolor{gray!10}
\textbf{QCSA + 11-keyframe submaps} & \checkmark & \checkmark & \checkmark & 19,271 & 1,602 & 31 & 98.06\% & \textbf{5.509}/\textbf{9.411} & \textbf{2} \\
\bottomrule
\end{tabular}
\vspace{2pt}

{\fontsize{6}{6.3}\selectfont
\begin{tabular}{@{}lrrrrrrrr@{}}
\toprule
\multicolumn{9}{l}{(b) ATE RMSE (m) by query$\to$database session and descriptor; entries are Top-1/QCSA} \\
\addlinespace[1pt]
Route & LiDAR Iris & ScanContext++ & NDT-MC & OT & LoGG3D-Net & RING++ & MinkLoc3Dv2 & Mean \\
\midrule
01/08 $\to$ 01/15 & $5.17/4.83$ & $7.43/3.81$ & $3.44/3.39$ & $3.64/23.55$ & $28.30/3.67$ & $7.84/3.76$ & $4.15/3.78$ & $8.57/\mathbf{6.68}$ \\
01/15 $\to$ 01/08 & $4.37/4.57$ & $3.62/4.13$ & $4.07/6.08$ & $3.91/14.91$ & $3.55/4.21$ & $4.52/3.97$ & $4.80/4.04$ & $\mathbf{4.12}/5.99$ \\
\addlinespace[1.2pt]
01/08 $\to$ 02/02 & $7.26/4.94$ & $7.87/2.78$ & $6.42/5.78$ & $10.59/4.38$ & $3.26/4.32$ & $5.20/7.65$ & $7.64/5.49$ & $6.89/\mathbf{5.05}$ \\
02/02 $\to$ 01/08 & $6.86/6.08$ & $7.47/7.69$ & $10.56/8.81$ & $12.15/9.38$ & $7.55/9.41$ & $10.36/7.30$ & $11.58/5.55$ & $9.51/\mathbf{7.75}$ \\
\addlinespace[1.2pt]
01/15 $\to$ 02/02 & $2.79/3.86$ & $2.70/3.02$ & $3.33/2.56$ & $6.15/3.99$ & $4.03/3.80$ & $2.39/2.43$ & $2.36/3.08$ & $3.39/\mathbf{3.25}$ \\
02/02 $\to$ 01/15 & $5.12/4.66$ & $4.02/3.48$ & $4.53/3.08$ & $10.88/7.17$ & $3.23/3.30$ & $3.10/3.32$ & $3.14/5.36$ & $4.86/\mathbf{4.34}$ \\
\bottomrule
\end{tabular}}
\end{table*}

Under these shared settings, QCSA sends 24.9\% fewer candidates to registration than Top-1. It reduces constraints exceeding the pose-error threshold by 35.4\%, lowers mean and 95th-percentile ATE by 11.5\% and 18.5\%, and reduces the number of configurations above 10\,m from seven to two. At the same time, it inserts 1,571 accurate constraints, 99.4\% as many as Top-1's 1,581.

The ablations separate hard-negative score calibration from threshold selection. Replacing the hard-negative LLR with normalized raw scores leaves registration attempts nearly unchanged but raises 95th-percentile ATE from $9.411$ to $14.601$\,m. Raw-score segments yield lower ATE in 23 configurations and QCSA in 19. QCSA's lower aggregate 95th-percentile ATE is concentrated in the high-error configurations.

For $\delta=0.20$--$0.40$, the 95th-percentile ATE is $9.36$--$9.56$\,m; at $\delta=0.02$, fewer segments are admitted and the 95th percentile rises to $34.81$\,m. The zero-threshold HN-LLR baseline uses 11.8\% fewer registrations and has four fewer pose-error constraints, with nearly the same 95th-percentile ATE\@. Query-level calibration admits 80 additional accurate constraints and lowers mean ATE by 1.9\%; it therefore adjusts verification coverage rather than relative-pose precision.

Table~\ref{tab:nclt_matched}(b) reports all 42 Top-1/QCSA ATE pairs. QCSA is lower in 25 configurations and on average for five of six routes; 01/15 $\to$ 01/08 is the exception. The two QCSA configurations above 10\,m both use OT on the January pair.

\subsection{NCLT Results with Segment Submaps}
Table~\ref{tab:nclt}(a) evaluates the complete QCSA front end, in which admitted segments form the registration submaps. It inserts 1,495 constraints, 1,476 within $0.5$\,m/$5^\circ$, and gives $6.42$\,m mean ATE\@. The corresponding single-scan configuration gives $25.39$\,m, demonstrating the importance of local-submap registration in the complete front end. Fixed 11-keyframe submaps give lower mean ATE ($5.509$\,m), whereas selected-segment submaps insert fewer constraints. Table~\ref{tab:nclt_matched} provides the controlled selection comparison. Because an interval is reserved only after successful registration, each registration-input condition is rerun chronologically and can alter later admissions.

\begin{table}[!b]
\centering
\caption{NCLT descriptor-based loop closure and MapClosures local-map association. In (a), only QCSA with segment submaps uses submap GICP.}
\label{tab:nclt}
\setlength{\tabcolsep}{1.50pt}
\scriptsize
\begin{tabular}{@{}lrrrr@{}}
\toprule
\multicolumn{5}{l}{(a) Descriptor-retrieval comparison} \\
Method & Constraints & \multicolumn{2}{c}{Precision (\%)} & ATE (m) \\
\cmidrule(lr){3-4}
& & Place & Pose & mean/max. \\
\midrule
Top-1 + scan GICP & 640 & 98.59 & 77.19 & 22.79 / 45.42 \\
Top-4 RANSAC & 636 & 99.37 & 74.84 & 24.74 / 49.54 \\
Top-4 SpectralGV & 647 & 98.92 & 75.27 & 25.01 / 49.66 \\
SeqSLAM & 235 & 98.72 & 84.68 & 30.42 / 51.00 \\
QCSA + single scan & 473 & 99.15 & 80.97 & 25.39 / 50.13 \\
\textbf{QCSA + segment submaps} & 1,495 & \textbf{99.40} & \textbf{98.73} & \textbf{6.42 / 19.40} \\
\midrule
\multicolumn{5}{l}{(b) MapClosures comparison} \\
Method & \shortstack{Mean\\constraints} & \multicolumn{2}{c}{Local-map association (\%)} & ATE (m) \\
\cmidrule(lr){3-4}
& & Precision & Recall & mean \\
\midrule
MapClosures & 138.0 & 82.20 & \textbf{50.45} & $7.263\!\pm\!0.047$ \\
QCSA + segment submaps & 35.6 & \textbf{98.35} & 12.66 & \textbf{6.421} \\
\bottomrule
\end{tabular}
\end{table}

Across descriptors, place precision is 98.51--100\%, relative-pose precision is 96.64--99.55\%, and mean ATE is $4.522$--$8.602$\,m; each physical session pair exceeds 98.5\% relative-pose precision.

Table~\ref{tab:nclt}(b) retains each front end's local-map construction but uses the same six directed pairs and pose graph. Under the MapClosures overlap labels, QCSA's 1,495 constraints represent 1,214 unique local-map pairs. Association precision and recall are computed for each of the 42 configurations and then averaged; MapClosures ATE is the mean $\pm$ sample standard deviation of three runs. At the evaluated operating points, QCSA produces fewer, higher-precision constraints and 11.6\% lower mean ATE, whereas MapClosures recovers substantially more associations.

\subsection{Repeated Outdoor Geometry}
Table~\ref{tab:transfer} evaluates Bridge map merging with the same centered 11-keyframe submap construction and downstream processing. A common registration setting is selected on the calibration pair and fixed on the test pair. Loop constraints initialize 27 of 28 QCSA graphs and all baseline graphs; no Iris registration passes the common gate for Bridge01-to-Bridge04. Counts and precision use all 28 evaluations. On the 27 graphs initialized by all methods, QCSA gives $6.346$\,m mean ATE, versus $6.452$\,m for Top-1 and $6.431$\,m for FPFH-RANSAC\@. The differences are small, but QCSA requires 17.6\% and 22.7\% fewer registration attempts. QCSA leads in place and $1$\,m/$5^\circ$ pose precision; Top-1 leads at 0.5\,m.

\begin{table}[!b]
\centering
\caption{HeLiPR Bridge map merging with common 11-keyframe submaps. Merged: accepted inter-session loop; ATE: 27 configurations merged by all methods.}
\label{tab:transfer}
\setlength{\tabcolsep}{1.05pt}
\scriptsize
\begin{tabular}{@{}lrrrrrr@{}}
\toprule
Method & \shortstack{Reg.\\attempts} & Constr. & Merged & \multicolumn{2}{c}{Precision (\%)} & ATE (m) \\
\cmidrule(lr){5-6}
& & & & Place & Pose 0.5/1\,m/$5^\circ$ & mean/95th \\
\midrule
Top-1 & 1,103 & 553 & 28/28 & 96.75 & \textbf{44.67}/89.87 & 6.452 / 8.764 \\
Top-4 FPFH-RANSAC & 1,176 & 476 & 28/28 & 94.12 & 38.87/89.71 & 6.431 / 8.775 \\
\rowcolor{gray!10}
\textbf{QCSA} & \textbf{909} & 526 & 27/28 & \textbf{98.10} & 37.45/\textbf{90.87} & \textbf{6.346 / 8.599} \\
\bottomrule
\end{tabular}
\end{table}

\subsection{Trajectory Interpretation and Limitations}
The QCSA--MapClosures and Bridge results show that constraint precision, coverage, and trajectory accuracy need not vary together. QCSA selects loop candidates, while submap construction and registration remain separate front-end components.

The multi-descriptor results demonstrate compatibility rather than independent trials because all descriptors on a route share the same trajectory. Outdoor evaluation is limited to three NCLT and two Bridge session pairs. Although the orientation-free SNULib labels include opposite-heading pairs, they do not form sustained positive segments; sustained opposite-direction and brief intersection-only revisits therefore remain untested.

QCSA assumes motion-selected keyframes and representative descriptor- and setting-specific calibration, and it abstains when the fitted threshold is non-finite. SNULib ATE uses an A-LOAM/HBA reference rather than independently surveyed ground truth. Candidate selection proceeds chronologically, whereas the reported PGO and ATE are computed after each trajectory is complete. Calibration transfer, alternative keyframe policies, and incremental or robust back ends remain to be evaluated.

\section{Conclusion}
\label{sec:conclusion}

QCSA selects LiDAR loop candidates by accumulating hard-negative-calibrated evidence over neighboring keyframes and comparing the strongest segment with a query-calibrated threshold. Each admitted segment forms two local submaps for one GICP registration. Across seven descriptors on SNULib, QCSA's ATE RMSE averages $0.1188$\,m with 53.8\% fewer registration attempts than Top-1. With identical downstream processing on NCLT, QCSA reduces registration attempts, constraints exceeding the pose-error tolerance, and trajectory error while retaining 99.4\% of Top-1's accurate constraints. Bridge shows fewer registration attempts and slightly lower trajectory error under identical outdoor settings. QCSA requires representative descriptor- and setting-specific calibration; sustained opposite-direction revisits and other keyframe policies remain to be evaluated. These results show that calibrated evidence over neighboring keyframes can improve loop-candidate selection in repetitive environments.

\bibliographystyle{IEEEtran}
\bibliography{reference}

@article{Cadena2016,
title={Past, present, and future of simultaneous localization and mapping: Toward the robust-perception age},
  author={Cadena, Cesar and Carlone, Luca and Carrillo, Henry and Latif, Yasir and Scaramuzza, Davide and Neira, Jos{\'e} and Reid, Ian and Leonard, John J},
  journal={IEEE Trans. Robot.},
  volume={32},
  number={6},
  pages={1309--1332},
  year={2016},
  doi={10.1109/TRO.2016.2624754}
}

@article{Jung2024HeLiPR,
  title={{HeLiPR}: Heterogeneous {LiDAR} dataset for inter-{LiDAR} place recognition under spatiotemporal variations},
  author={Jung, Minwoo and Yang, Wooseong and Lee, Dongjae and Gil, Hyeonjae and Kim, Giseop and Kim, Ayoung},
  journal={Int. J. Robot. Res.},
  volume={43},
  number={12},
  pages={1867--1883},
  year={2024},
  doi={10.1177/02783649241242136}
}

@ARTICLE{HBA,
  title={Large-scale {LiDAR} consistent mapping using hierarchical {LiDAR} bundle adjustment},
  author={Liu, Xiyuan and Liu, Zheng and Kong, Fanze and Zhang, Fu},
  journal={IEEE Robot. Autom. Lett.},
  volume={8},
  number={3},
  pages={1523--1530},
  year={2023},
  doi={10.1109/LRA.2023.3238902}
}

@inproceedings{Kim2018ScanContext,
  title={{Scan Context}: Egocentric spatial descriptor for place recognition within {3D} point cloud map},
  author={Kim, Giseop and Kim, Ayoung},
  booktitle={Proc. IEEE/RSJ Int. Conf. Intell. Robots Syst. (IROS)},
  pages={4802--4809},
  year={2018},
  doi={10.1109/IROS.2018.8593953}
}

@inproceedings{Wang2020LiDAR-Iris,
  title={{LiDAR Iris} for loop-closure detection},
  author={Wang, Ying and Sun, Zezhou and Xu, Cheng-Zhong and Sarma, Sanjay E and Yang, Jian and Kong, Hui},
  booktitle={Proc. IEEE/RSJ Int. Conf. Intell. Robots Syst. (IROS)},
  pages={5769--5775},
  year={2020},
  doi={10.1109/IROS45743.2020.9341010}
}

@article{Kim2022ScanContext++,
  title={{Scan Context++}: Structural place recognition robust to rotation and lateral variations in urban environments},
  author={Kim, Giseop and Choi, Sunwook and Kim, Ayoung},
  journal={IEEE Trans. Robot.},
  volume={38},
  number={3},
  pages={1856--1874},
  year={2022},
  doi={10.1109/TRO.2021.3116424}
}

@article{Xu2023RING++,
  title={{RING++}: Roto-translation invariant gram for global localization on a sparse scan map},
  author={Xu, Xuecheng and Lu, Sha and Wu, Jun and Lu, Haojian and Zhu, Qiuguo and Liao, Yiyi and Xiong, Rong and Wang, Yue},
  journal={IEEE Trans. Robot.},
  volume={39},
  number={6},
  pages={4616--4635},
  year={2023},
  doi={10.1109/TRO.2023.3303035}
}

@inproceedings{Liao2024NDTMapCode,
  title={{NDT-Map-Code}: A {3D} global descriptor for real-time loop closure detection in {LiDAR} {SLAM}},
  author={Liao, Lizhou and Yan, Wenlei and Sun, Li and Bai, Xinhui and You, Zhenxing and Yuan, Hongyuan and Fu, Chunyun},
  booktitle={Proc. IEEE/RSJ Int. Conf. Intell. Robots Syst. (IROS)},
  pages={7854--7861},
  year={2024},
  doi={10.1109/IROS58592.2024.10801384}
}

@article{Gupta2026DensityMaps,
  title={Efficiently closing loops in {LiDAR}-based {SLAM} using point cloud density maps},
  author={Gupta, Saurabh and Guadagnino, Tiziano and Mersch, Benedikt and Trekel, Niklas and Malladi, Meher V. R. and Stachniss, Cyrill},
  journal={Int. J. Robot. Res.},
  year={2026},
  note={{OnlineFirst}},
  doi={10.1177/02783649261449269}
}

@inproceedings{Dube2017SegMatch,
  title={{SegMatch}: Segment based place recognition in {3D} point clouds},
  author={Dub{\'e}, Renaud and Dugas, Daniel and Stumm, Elena and Nieto, Juan and Siegwart, Roland and Cadena, Cesar},
  booktitle={Proc. IEEE Int. Conf. Robot. Autom. (ICRA)},
  pages={5266--5272},
  year={2017},
  doi={10.1109/ICRA.2017.7989618}
}

@inproceedings{Vidanapathirana2022LoGG3DNet,
  title={{LoGG3D-Net}: Locally guided global descriptor learning for {3D} place recognition},
  author={Vidanapathirana, Kavisha and Ramezani, Milad and Moghadam, Peyman and Sridharan, Sridha and Fookes, Clinton},
  booktitle={Proc. IEEE Int. Conf. Robot. Autom. (ICRA)},
  pages={2215--2221},
  year={2022},
  doi={10.1109/ICRA46639.2022.9811753}
}

@inproceedings{Komorowski2022MinkLoc3Dv2,
  title={Improving point cloud based place recognition with ranking-based loss and large batch training},
  author={Komorowski, Jacek},
  booktitle={Proc. Int. Conf. Pattern Recognit. (ICPR)},
  pages={3699--3705},
  year={2022},
  doi={10.1109/ICPR56361.2022.9956458}
}

@article{Ma2022OverlapTransformer,
  title={{OverlapTransformer}: An efficient and yaw-angle-invariant transformer network for {LiDAR}-based place recognition},
  author={Ma, Junyi and Zhang, Jun and Xu, Jintao and Ai, Rui and Gu, Weihao and Chen, Xieyuanli},
  journal={IEEE Robot. Autom. Lett.},
  volume={7},
  number={3},
  pages={6958--6965},
  year={2022},
  doi={10.1109/LRA.2022.3178797}
}

@article{Ma2023SeqOT,
  title={{SeqOT}: A spatial--temporal transformer network for place recognition using sequential {LiDAR} data},
  author={Ma, Junyi and Chen, Xieyuanli and Xu, Jingyi and Xiong, Guangming},
  journal={IEEE Trans. Ind. Electron.},
  volume={70},
  number={8},
  pages={8225--8234},
  year={2023},
  doi={10.1109/TIE.2022.3229385}
}

@inproceedings{MilfordWyeth2012SeqSLAM,
  title={{SeqSLAM}: Visual route-based navigation for sunny summer days and stormy winter nights},
  author={Milford, Michael J and Wyeth, Gordon F},
  booktitle={Proc. IEEE Int. Conf. Robot. Autom. (ICRA)},
  pages={1643--1649},
  year={2012},
  doi={10.1109/ICRA.2012.6224623}
}

@inproceedings{Vysotska2025AdaptiveThresholding,
  title={Adaptive thresholding for sequence-based place recognition},
  author={Vysotska, Olga and Bogoslavskyi, Igor and Hutter, Marco and Stachniss, Cyrill},
  booktitle={Proc. IEEE Int. Conf. Robot. Autom. (ICRA)},
  pages={2219--2225},
  year={2025},
  doi={10.1109/ICRA55743.2025.11128422}
}

@inproceedings{Vidanapathirana2021Locus,
  title={{Locus}: {LiDAR}-based place recognition using spatiotemporal higher-order pooling},
  author={Vidanapathirana, Kavisha and Moghadam, Peyman and Harwood, Ben and Zhao, Muming and Sridharan, Sridha and Fookes, Clinton},
  booktitle={Proc. IEEE Int. Conf. Robot. Autom. (ICRA)},
  pages={5075--5081},
  year={2021},
  doi={10.1109/ICRA48506.2021.9560915}
}

@article{Cummins2008FABMAP,
  title={{FAB-MAP}: Probabilistic localization and mapping in the space of appearance},
  author={Cummins, Mark and Newman, Paul},
  journal={Int. J. Robot. Res.},
  volume={27},
  number={6},
  pages={647--665},
  year={2008},
  doi={10.1177/0278364908090961}
}

@inproceedings{Sunderhauf2012Switchable,
  title={Switchable constraints for robust pose graph {SLAM}},
  author={S{\"u}nderhauf, Niko and Protzel, Peter},
  booktitle={Proc. IEEE/RSJ Int. Conf. Intell. Robots Syst. (IROS)},
  pages={1879--1884},
  year={2012},
  doi={10.1109/IROS.2012.6385590}
}

@inproceedings{Latif-RSS-12,
  title={Robust loop closing over time},
  author={Latif, Yasir and Cadena Lerma, Cesar and Neira, Jos{\'e}},
  booktitle={Proc. Robot.: Sci. Syst. (RSS)},
  address={Sydney, Australia},
  month={July},
  year={2012},
  doi={10.15607/RSS.2012.VIII.030}
}

@inproceedings{zhang2014loam,
  title={{LOAM}: Lidar odometry and mapping in real-time},
  author={Zhang, Ji and Singh, Sanjiv},
  booktitle={Proc. Robot.: Sci. Syst. (RSS)},
  address={Berkeley, CA, USA},
  month={July},
  year={2014},
  doi={10.15607/RSS.2014.X.007}
}

@misc{Qin2018ALOAM,
  title={{A-LOAM}: Advanced implementation of {LOAM}},
  author={Qin, Tong and Cao, Shaozu},
  year={2019},
  note={[Online]. Available: \url{https://github.com/HKUST-Aerial-Robotics/A-LOAM}}
}

@article{angelopoulos2023conformal,
  title={Conformal prediction: A gentle introduction},
  author={Angelopoulos, Anastasios N and Bates, Stephen},
  journal={Found. Trends Mach. Learn.},
  volume={16},
  number={4},
  pages={494--591},
  year={2023},
  doi={10.1561/2200000101}
}

@article{nclt,
  title     = {{University of Michigan North Campus Long-Term Vision and LiDAR Dataset}},
  author    = {Carlevaris-Bianco, Nicholas and Ushani, Arash K. and Eustice, Ryan M.},
  journal   = {Int. J. Robot. Res.},
  volume    = {35},
  number    = {9},
  pages     = {1023--1035},
  year      = {2016},
  doi       = {10.1177/0278364915614638}
}

@article{FischlerBolles1981,
  author  = {Fischler, Martin A. and Bolles, Robert C.},
  title   = {Random Sample Consensus: A Paradigm for Model Fitting with Applications to Image Analysis and Automated Cartography},
  journal = {Commun. ACM},
  volume  = {24},
  number  = {6},
  pages   = {381--395},
  year    = {1981},
  doi     = {10.1145/358669.358692}
}

@inproceedings{Rusu2009FPFH,
  author    = {Rusu, Radu Bogdan and Blodow, Nico and Beetz, Michael},
  title     = {Fast Point Feature Histograms ({FPFH}) for {3D} Registration},
  booktitle = {Proc. IEEE Int. Conf. Robot. Autom. (ICRA)},
  pages     = {3212--3217},
  year      = {2009},
  doi       = {10.1109/ROBOT.2009.5152473}
}

@article{Vidanapathirana2023SpectralGV,
  author  = {Vidanapathirana, Kavisha and Moghadam, Peyman and Sridharan, Sridha and Fookes, Clinton},
  title   = {Spectral Geometric Verification: Re-Ranking Point Cloud Retrieval for Metric Localization},
  journal = {IEEE Robot. Autom. Lett.},
  volume  = {8},
  number  = {5},
  pages   = {2494--2501},
  year    = {2023},
  doi     = {10.1109/LRA.2023.3255560}
}

@article{Denniston2022Prioritization,
  author  = {Denniston, Christopher E. and Chang, Yun and Reinke, Andrzej and Ebadi, Kamak and Sukhatme, Gaurav S. and Carlone, Luca and Morrell, Benjamin and Agha-mohammadi, Ali-akbar},
  title   = {Loop Closure Prioritization for Efficient and Scalable Multi-Robot {SLAM}},
  journal = {IEEE Robot. Autom. Lett.},
  volume  = {7},
  number  = {4},
  pages   = {9651--9658},
  year    = {2022},
  doi     = {10.1109/LRA.2022.3191156}
}

@inproceedings{Laserna2025CliReg,
  title={Loop closure via maximal cliques in {3D} {LiDAR}-based {SLAM}},
  author={Laserna, Javier and Gupta, Saurabh and Martinez Mozos, Oscar and Stachniss, Cyrill and San Segundo, Pablo},
  booktitle={Proc. Eur. Conf. Mobile Robots (ECMR)},
  pages={1--6},
  year={2025},
  doi={10.1109/ECMR65884.2025.11163179}
}

@inproceedings{Segal2009GICP,
  author    = {Segal, Aleksandr and H{\"a}hnel, Dirk and Thrun, Sebastian},
  title     = {Generalized-{ICP}},
  booktitle = {Proc. Robot.: Sci. Syst. (RSS)},
  address   = {Seattle, WA, USA},
  month     = {June},
  year      = {2009},
  doi       = {10.15607/RSS.2009.V.021}
}

@inproceedings{Shan2020LIOSAM,
  author    = {Shan, Tixiao and Englot, Brendan and Meyers, Drew and Wang, Wei and Ratti, Carlo and Rus, Daniela},
  title     = {{LIO-SAM}: Tightly-coupled Lidar inertial odometry via smoothing and mapping},
  booktitle = {Proc. IEEE/RSJ Int. Conf. Intell. Robots Syst. (IROS)},
  pages     = {5135--5142},
  year      = {2020},
  doi       = {10.1109/IROS45743.2020.9341176}
}

@book{Silverman1986,
  author    = {Silverman, Bernard W.},
  title     = {Density Estimation for Statistics and Data Analysis},
  publisher = {Chapman and Hall},
  address   = {London, UK},
  year      = {1986}
}
\end{document}